\documentclass[letterpaper]{article} 
\usepackage{aaai24}  
\usepackage{times}  
\usepackage{helvet}  
\usepackage{courier}  
\usepackage[hyphens]{url}  
\usepackage{graphicx} 
\usepackage{natbib}  
\usepackage{caption} 
\usepackage{algorithm}
\usepackage{algorithmic}

\usepackage{newfloat}
\usepackage{listings}
\DeclareCaptionStyle{ruled}{labelfont=normalfont,labelsep=colon,strut=off} 
\floatstyle{ruled}
\newfloat{listing}{tb}{lst}{}
\floatname{listing}{Listing}
\usepackage{amssymb}
\usepackage{amsmath}
\usepackage{booktabs}
\usepackage{mathtools}
\usepackage{multirow}
\usepackage{graphicx}
\usepackage{cleveref}
\usepackage{svg}
\usepackage{subcaption}

\title{Redefining the Laparoscopic Spatial Sense: AI-based Intra- and Postoperative Measurement from Stereoimages}
\author{
    Leopold Müller\textsuperscript{\rm 2,3,4},  
    Patrick Hemmer\textsuperscript{\rm 3},
    Moritz Queisner\textsuperscript{\rm 1},
    Igor Sauer\textsuperscript{\rm 1},
    Simeon Allmendinger\textsuperscript{\rm 2,3,4},
    Johannes Jakubik\textsuperscript{\rm 3},
    Michael Vössing\textsuperscript{\rm 3},
    Niklas Kühl\textsuperscript{\rm 2,3,4},\\
}
\affiliations{
    \textsuperscript{\rm 1}Charité Universitätsmedizin Berlin,
    \textsuperscript{\rm 2}Fraunhofer FIT,
    \textsuperscript{\rm 3}Karlsruhe Institute of Technology,
    \textsuperscript{\rm 4}University of Bayreuth\\

    leopold.mueller@uni-bayreuth.de
}

\begin{document}

\maketitle

\begin{abstract}
A significant challenge in image-guided surgery is the accurate measurement task of relevant structures such as vessel segments, resection margins, or bowel lengths. While this task is an essential component of many surgeries, it involves substantial human effort and is prone to inaccuracies. In this paper, we develop a novel human-AI-based method for laparoscopic measurements utilizing stereo vision that has been guided by practicing surgeons. Based on a holistic qualitative requirements analysis, this work proposes a comprehensive measurement method, which comprises state-of-the-art machine learning architectures, such as RAFT-Stereo and YOLOv8. The developed method is assessed in various realistic experimental evaluation environments. Our results outline the potential of our method achieving high accuracies in distance measurements with errors below 1 mm. Furthermore, on-surface measurements demonstrate robustness when applied in challenging environments with textureless regions. Overall, by addressing the inherent challenges of image-guided surgery, we lay the foundation for a more robust and accurate solution for intra- and postoperative measurements, enabling more precise, safe, and efficient surgical procedures.
\end{abstract}

\section{Introduction}

In surgical procedures, accurate measurements are crucial. The determination of precise arterial diameters directs the choice of endovascular prostheses \cite{carvalho_measuring_2006}. When a tumour is removed, the optimal resection margin must be selected \cite{Bilgeri2020TheEO}. This ensures complete removal of the tumour while sparing the surrounding healthy tissue. Similarly, precise measurements are necessary during bowel surgeries to preserve functional bowel length and ensure optimal nutrient absorption after the operation \cite{Gazer2017AccuracyAI}. In open surgery, direct visualization and tactile feedback facilitate accurate measurements of tissue and anatomical structures. However, laparoscopic surgery introduces two inherent challenges in intraoperative measurements. First, unlike open surgery, structures in laparoscopic procedures cannot be measured directly due to the limited working space and the use of specialized instruments, rendering traditional measurements exceedingly difficult or even impossible. Second, indirect perception of structures via cameras introduces limited depth perception and distorted perspective, making it challenging to accurately estimate size relationships and obtain reliable measurements. These two factors highlight the necessity of a measurement tool specifically tailored for laparoscopic procedures.

In this study, we introduce a versatile measurement method for laparoscopic surgery, aiming to provide a solution that can measure any structure, both intra- and postoperatively. This leads to the following benefits: It simplifies the measurement process by being applicable to a range of use cases. A readily available measurement method can support the decision-making process during surgery and enable more comprehensive, quantified quality assurance postoperatively. In addition, increased measurements contribute to a broader data foundation, allowing for the identification of best practices and integration into intelligent systems, such as automatic surgery reports. While the method is fundamentally designed for minimally invasive procedures, we focus on laparoscopy as a prime example.

Our contribution to the field of laparoscopic surgery is threefold. First, in collaboration with medical experts, we identify key requirements for a laparoscopic measurement method. Second, we develop a measurement method that meets these key requirements and implement this proposed method using state-of-the-art components from the realm of computer vision. Third, we evaluate its performance through various qualitative and quantitative experiments, demonstrating that the method is not only highly accurate but also robust in challenging conditions such as textureless regions, blood, reflections, and smoke. The code is made available\footnote{\url{https://github.com/leopoldmueller/LaparoscopicMeasurement}}.

\section{Related Work}

Our work resides at the intersection of two primary areas of research: surface reconstruction methods used in laparoscopy and methods of obtaining precise measurements in minimally invasive surgery (MIS). In the following, we briefly outline related work in both areas.

\subsection{Surface Reconstruction in Surgery}

In the realm of laparoscopy, surface reconstruction is an essential technique aimed at compensating for challenges like a restricted field of view, complex hand-eye coordination, and absence of tactile feedback \cite{maier-hein_comparative_2014}. Many approaches have been developed to create 3D models of organic surfaces that can facilitate data registration and augmented reality in laparoscopic procedures \cite{rohl_dense_2012, maier-hein_comparative_2014}.

Surface reconstruction methods can be categorized into two types: active and passive methods. Active methods, such as structured light and Time of Flight (ToF), necessitate controlled light projection into the environment and have been explored in multiple studies \cite{ackerman_surface_2002, maurice_structured_2012, maier-hein_comparative_2014}. Conversely, passive methods do not require projected light and rely solely on camera images. Various algorithms have been proposed for passive methods, such as stereo reconstruction and photometric stereo \cite{kittler_real-time_2013, hutchison_3d_2013, rohl_real-time_2011, hutchison_3d_2012, malti_combining_2014}.

Among these methods, comparative studies have shown the passive stereo-based approaches to be more advantageous due to their higher accuracy and dense point clouds, in comparison to methods like ToF \cite{maier-hein_comparative_2014, groch_3d_2011}.

While our method also performs a surface reconstruction to capture accurate 3D distances, in distinction to existing work, our objective differs from registering 3D models. Instead, our goal is to precisely measure structures in laparoscopy. Based on the results of the method comparison, we opt for a stereo-based method. In addition to the advantages of more accurate results and dense point clouds, which allow for finer measurements, stereo vision offers another crucial benefit. As a passive system, it relies solely on camera images. This is a significant criterion given the limited working area in laparoscopy and enables postoperative measurement. Furthermore, by utilizing stereo vision, we rely on a component that addresses limited depth perception in laparoscopy \cite{way_causes_2003}.

Certain methods, such as Intuitive Surgical's da Vinci System, have employed stereo vision to optimize 3D perception. Stereo vision has already been used as a component in laparoscopy for various reasons that are beyond the scope of this work. Consequently, it is reasonable to capitalize on this existing technology and develop further functionalities using the available components.

\subsection{Measurement Approaches}

Despite the advancements in MIS, accurate intraoperative measurement remains a challenging task. Some current approaches demonstrate the potential of stereo-endoscopes in obtaining precise measurements. For instance, \citet{field_stereo_2009} demonstrated that accurate anatomical measurements could be achieved using stereo-endoscopes. However, their method relied on markers, limiting its applicability. Similarly, \citet{bodenstedt_image-based_2016} presented an image-based approach for bowel measurement in laparoscopy using stereo endoscopy. Despite showing promise in phantom and porcine datasets, this method was limited by the need for previous images for reconstruction and its specificity to certain structures.

Our goal is to extend this concept and provide a general-purpose measurement solution that can be used for any structure without the need for manually placed markers or previous images. To accomplish this, we propose using state-of-the-art AI-based disparity estimation for surface reconstruction \cite{lipson_raft-stereo_2021}.

\begin{figure*}[htbp]
    \centering
    \includegraphics[width=.89\textwidth]{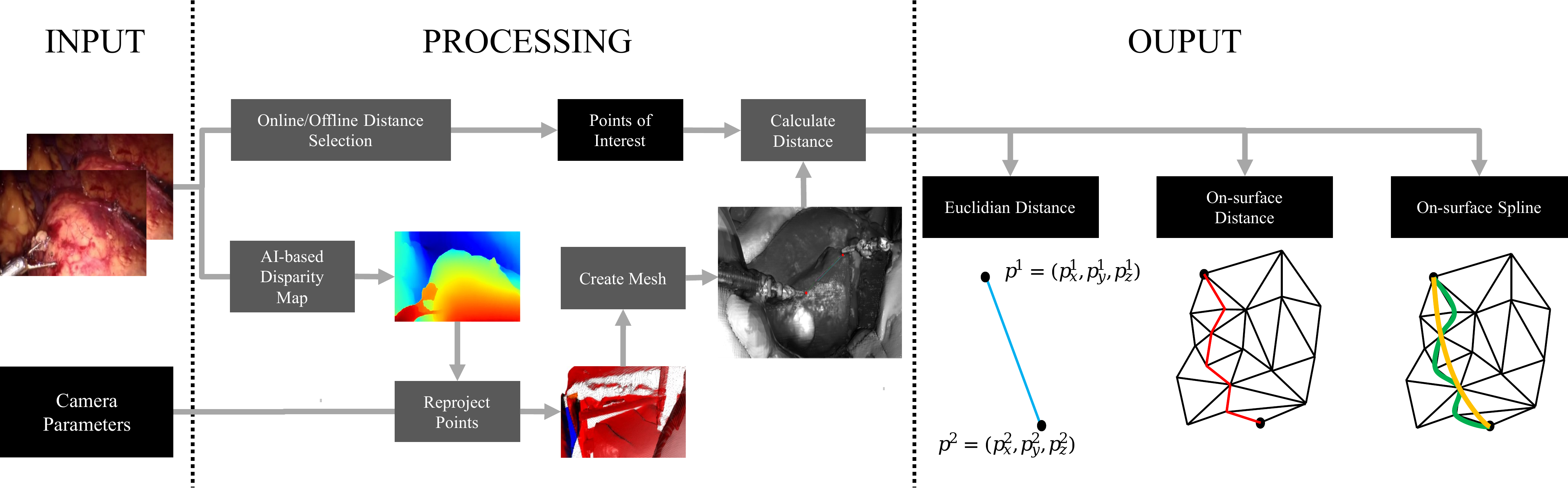}
    \caption{Schematic illustration of the measurement method divided into input, processing and output.}
    \label{fig:method}
\end{figure*}

\section{Method Requirements}
\label{sec:keyrequirements}

We formulate the requirements after a thorough review of the characteristics of laparoscopic surgery and in collaboration with medical experts at Charité Universitätsmedizin Berlin, which included the analysis of an actual surgery and analyses of potential case studies. The inclusion of these requirements in the current study aims to maximize the benefits of laparoscopic surgery and address its challenges.

\begin{enumerate}

\item[R1] \textbf{Utilize existing components:} When performing a laparoscopic surgery the measurement tool should not require additional instruments or incisions, as one of the primary benefits of laparoscopic surgery is its minimally invasive nature. The solution should rely on existing components and tools that are already used in the surgical environment.

\item[R2] \textbf{Integrate as passive system:} When surgeons want to perform a measurement, the system must seamlessly integrate without altering the surgeon's workflow and the surgical procedure should remain the primary focus.

\item[R3] \textbf{Increase ease of use and availability:} When a surgeon needs to access the measurement method during surgery, then the system should be easily usable and available without preparation. This facilitates better documentation, data collection, and decision-making without interrupting the surgical process or requiring preparation before measurements are taken.

\item[R4] \textbf{Leverage existing camera input:} When processing visual data for the surgery, then the system should utilize the input stream from the laparoscopic camera, ensuring it aligns with what the surgeon sees during the procedure.

\item[R5] \textbf{Implement integration of computer vision-based approaches:} When integrating with existing approaches like phase recognition, then the system must employ the same data sources to ensure compatibility and seamless integration with these methods.

\item[R6] \textbf{Adhere to medical standards:} When operating in the sensitive surgical domain where patient safety is crucial, then the system must ensure high accuracy and robustness in all measurements.

\item[R7] \textbf{Enhance versatility of measurement tasks:} When a surgeon aims to measure a structure during surgery, the system should be capable of measuring any structure of interest, eliminating the need for multiple specialized tools tailored to specific tasks.

\item[R8] \textbf{Provide intra- and postoperative measurement capabilities:} When taking measurements during a surgical procedure, then the system should support both intra- and postoperative evaluations as they can provide valuable data for further research, quality control, and outcome analysis, contributing to the continuous improvement of surgical practices.

\end{enumerate}

\section{Method}

Our method depicted in \Cref{fig:method} is structured into three parts: Input, Processing, and Output.


\paragraph{Input.} To perform a measurement using our framework, two pieces of information are required: Stereo images that contain the structure to be measured and the camera parameters of the stereo camera. The stereo images are used to select the desired distance, while the camera parameters are used to project the 2D image into 3D space. While we obtain the images directly from the camera's input stream, the camera parameters must be determined at the outset through camera calibration.

\paragraph{Processing.} We define the left image $I_L$ as our source image to which we relate all measurements. We assume that the image shows all points of the real world discretized as pixels in the image. If we want to measure the distance between two points of the original scene in world coordinates $p_a^{W}$ and $p_b^{W}$, we need to select two pixels in the image $p_a^{I}$ and $p_b^{I}$ containing these points. For the selection of these two points, we distinguish between two cases: offline and online selection.

The offline selection represents the option that a person is using an interface where the points of interest are selected manually by clicking on the specific pixels in the left image $I_L$ with a controller, e.g. mouse or touchscreen. This case offers the possibility to measure postoperative on the video or intraoperative with the help of an assistant that performs the task.

To enable a simple and non-intrusive measurement process that supports the medical workflow by continuously offering the possibility of taking measurements for both documentation and decision-making purposes, we introduce the online selection functionality. With this, surgeons can directly use the available surgical instruments, such as two graspers, as measuring tools during the surgery. The surgeon first positions the tools at both ends of the structure to be measured, ensuring that the tooltips point towards $p_a^{I}$ and $p_b^{I}$. This allows the surgeon to define the structure to be measured within the scene. Then the tips of the instruments are automatically detected and set as $p_a^{I}$ and $p_b^{I}$. To achieve this, We define a function $\sigma: I_L \rightarrow (M_a, M_b)$, where $M_a$ and $M_b$ are the segmentation masks for surgical tools a and b, respectively. We propose YOLOv8 \cite{jocher_yolov8_2023} as segmentation model $\sigma$. The output masks have binary values, with 1 representing the presence of a tool and 0 indicating no tool at the corresponding pixel location. Similar to \citet{bodenstedt_image-based_2016}, we calculate the center of mass $m_a$ and $m_b$ for the masks $M_a$ and $M_b$, respectively, by finding the average coordinates of all non-zero elements in each mask. The points farthest from their respective centroids are then identified as $p_a^{I}$ and $p_b^{I}$.
 
Given the two points $p_a^{I}$ and $p_b^{I}$ in pixel coordinates, we must first reproject them back to world coordinates to measure the distance between our points. To do this, we start with the AI-based estimation of the disparity map using RAFT-Stereo \cite{lipson_raft-stereo_2021}. RAFT-Stereo is a deep learning architecture for rectified stereo, building upon the RAFT optical flow network \cite{Teed2020}. It introduces multi-level convolutional gated recurrent units to more efficiently propagate information across images. Then we derive the reprojection matrix $Q$ from the camera parameters and transform the pixels back to world coordinates. Given a reprojected point cloud $P = \{p_1^{W}, p_2^{W}, \ldots, p_n^{W}\}$, where $p_i^{W} \in \mathbb{R}^3$, obtained from a reprojected image, we aim to reconstruct a smooth, watertight mesh of the underlying surface consisting of well defined triangles. To do this, we use the Poisson Surface Reconstruction (PSR) algorithm \cite{kazhdan_poisson_2006}.

\paragraph{Output.} Our method computes two different types of distances: the direct and the on-surface distance between points $p_a^{W}$ and $p_b^{W}$. The direct distance is the Euclidean distance between the selected points (\Cref{fig:method} blue line). Considering the complex structure of human anatomy, we include an on-surface distance measurement (\Cref{fig:method} red line). From the triangulated mesh representation of the anatomical structure, we create an undirected graph $G = (V, E)$, where $V$ is the set of vertices and $E$ is the set of edges. Each vertex in $V$ corresponds to a vertex in the mesh, and each edge in $E$ connects two vertices that share a face in the mesh. The weight of each edge $(i, j)$ is defined as the Euclidean distance between the responding vertices. We compute the optimal path distance between vertices belonging to \(p_a^{W}\) and \(p_b^{W}\) using Dijkstra's shortest path algorithm \cite{bodenstedt_image-based_2016}. To address the potential overestimation of distance due to the vertex structure in the mesh, we use spline interpolation on the path points \cite{bodenstedt_image-based_2016} (\Cref{fig:method} green and yellow line).

\section{Experiments}
\label{sec:experimentalsetup}

In this section, we first introduce all system components of our prototype implementation of the method and then elaborate on the evaluation of the conducted experiments.

\subsection{Experimental Setup}

We first provide an overview of the system components that are common to all six experiments.

\paragraph{Stereo Camera.} For our experimental setup, we utilize the Intel RealSense D435 camera as a stereo camera to simulate a laparoscope. The camera is equipped with two infrared sensors, which provide the necessary stereo input for the measurement system. We use a resolution of 848x480 with a frame rate of 15 fps. The camera captures monochrome images, which presents an additional challenge for the measurement system's robustness since they contain less information than RGB images.

\paragraph{Segmentation Model.} For real-time measurements, we draw upon YOLOv8 \cite{jocher_yolov8_2023} as segmentation model. We train the model using the Ex-vivo dVRK segmentation dataset related to the work of Colleoni, Edwards and Stoyanov \shortcite{colleoni_synthetic_2020}. Since the Intel RealSense camera provides monochrome stereo input streams, we convert the RGB images to monochrome. The data is split into 70\% for training, 20\% for validation, and 10\% for testing.

\paragraph{Disparity Algorithm.} We use two different algorithms. The first one is RAFT-Stereo, which serves as a component for disparity estimation in our method. Due to the high generalization of RAFT-Stereo compared to related work in the field, we use a model pre-trained on a Middlebury Stereo dataset \cite{lipson_raft-stereo_2021}. We believe that having an accurately trained model is crucial, rather than relying on poor training data within the domain. The second is the well-known SGBM \cite{hirschmuller_accurate_2005} as a baseline to which we benchmark our framework, since related literature does not provide code for a general comparison of the approaches.


\paragraph{Surface reconstruction method.} In our experimental setup, we utilize the PSR method from the Open3D library to reconstruct 3D surfaces \cite{zhou_open3d_2018}.

        

\begin{figure}[ht]%
\centering
\setlength{\abovecaptionskip}{0pt} 
    \begin{tabular}{p{2.4cm} p{2.4cm} p{2.4cm}}
        \includegraphics[width=2.4cm]{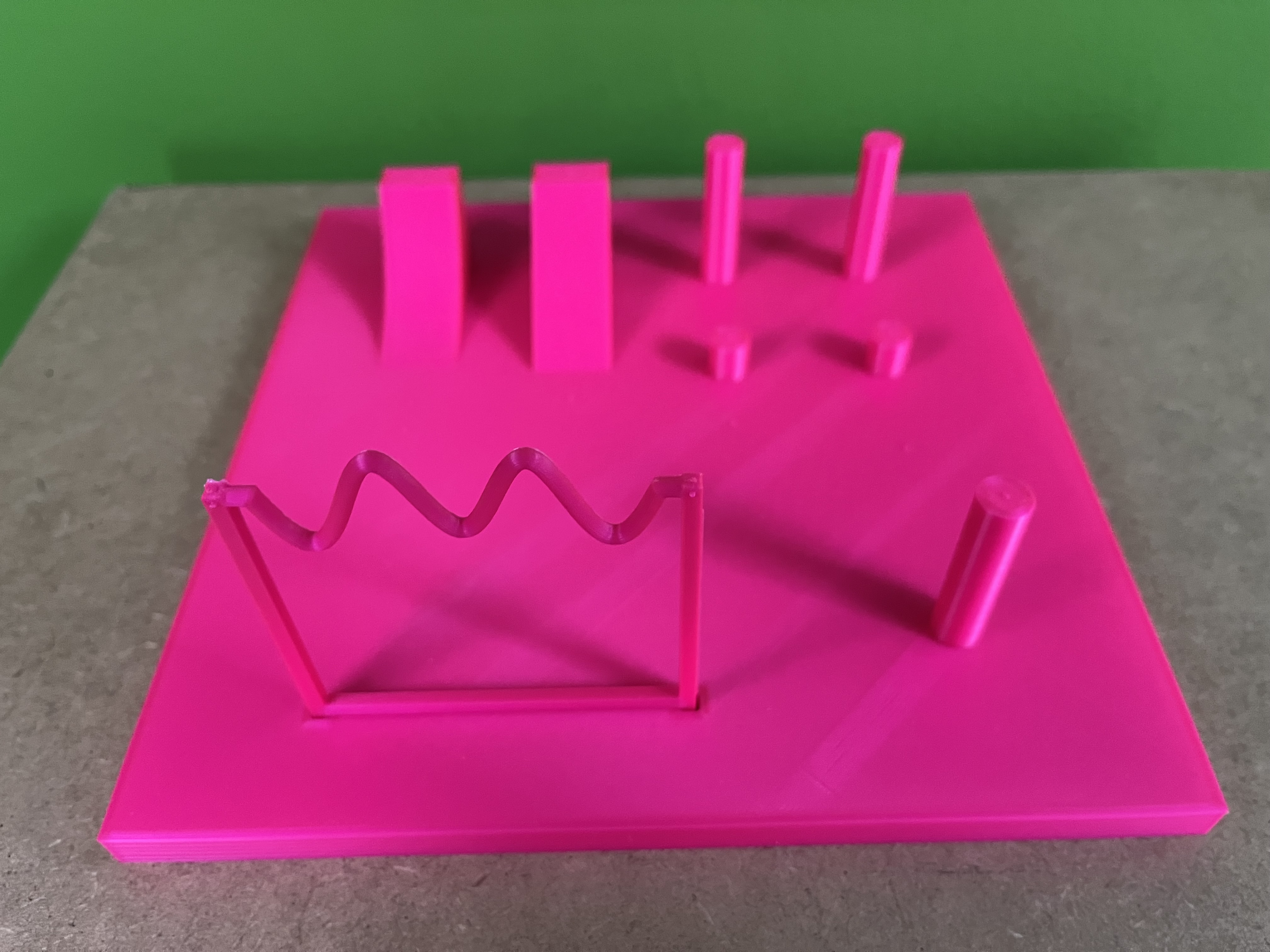} &
        \includegraphics[width=2.4cm]{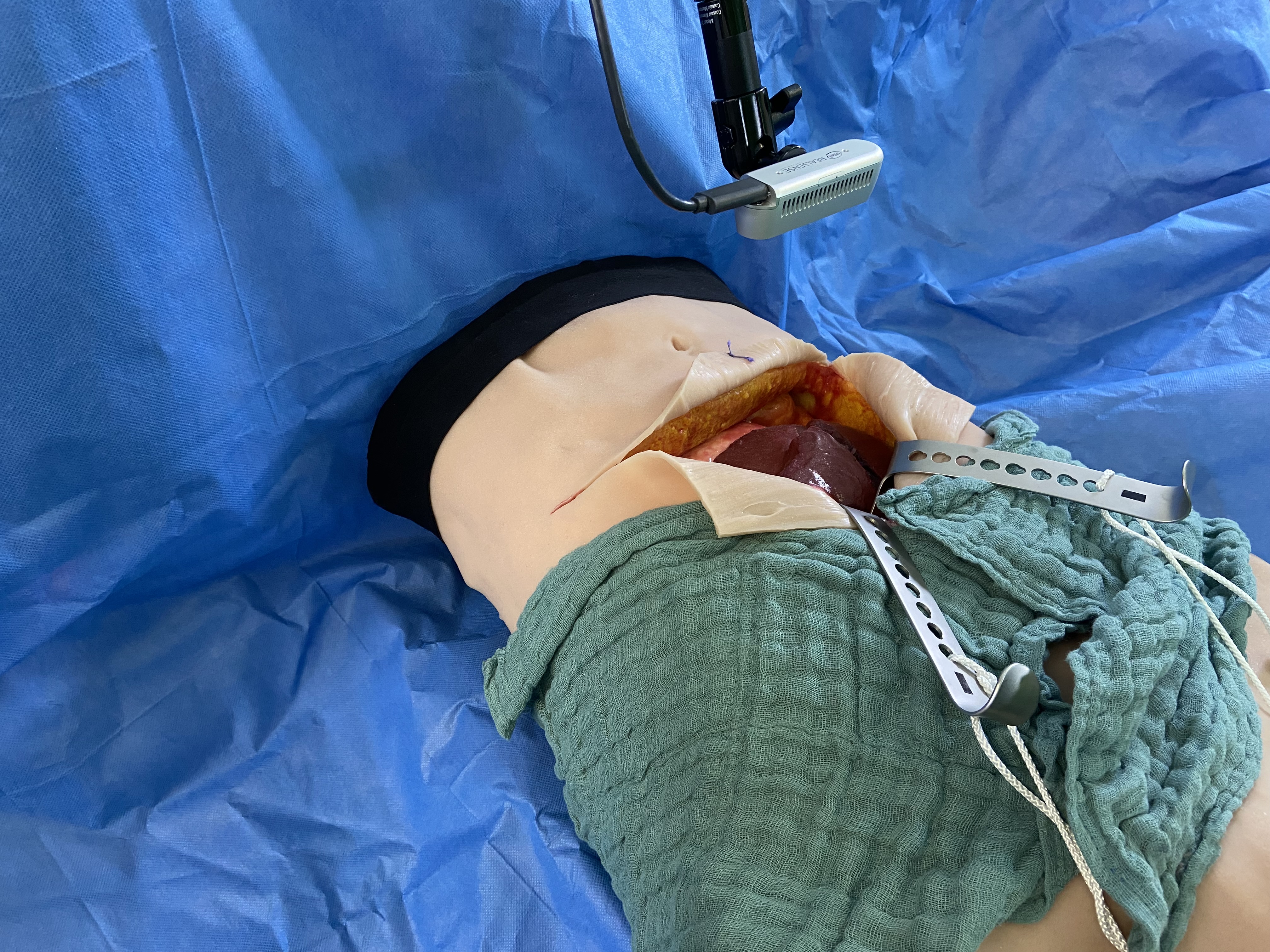} &
        \includegraphics[width=2.4cm]{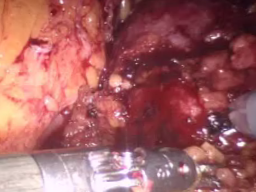} \\
        \centering \small Playground. &
        \centering \small Surgical Phantom. &
        \centering \small Real surgery \cite{ye_self-supervised_2017}. \\
        
    \end{tabular}
    \caption{Environments used in the experiments.}
    \label{fig:environments}
\end{figure}

\subsection{Evaluation}

To evaluate the proposed method, we deploy it in three environments, each designed to assess different functionalities.

\paragraph{Quantitative Evaluation.} Evaluating the accuracy of the method requires an environment with well-known structures for measurement. In collaboration with medical experts, we designed a CAD model that abstracts and simplifies the most critical real-world structures. We refer to this environment as the ``Playground'' (\Cref{fig:environments}). The CAD model was 3D printed using PLA material with a layer thickness of 0.1 mm. Reference points are consistently marked with a small protrusion, enabling their manual recognition in the stereo images. The reference points are applied to different structures in such a way that certain distances occur multiple times in different orientations and positions. To evaluate the on-surface measurements, plane, convex, and concave shapes were incorporated into the Playground in addition to the direct distances. The measurements should not only be highly accurate but also deliver robust measurement results over several trials. The design of the Playground and the measurements taken from various perspectives serve to control this property. The Playground is printed in a single color with a smooth surface, resulting in a structureless surface prone to strong reflections. Combined, this represents an extreme case for stereo vision.

We perform multiple measurements of known distances and analyze the error, mean absolute error (MAE), and the standard deviation (STD) of the error: We conduct $n=48$ offline and online measurements for the ground truth distances $l_{\text{gt, direct}} = \{40\, \text{mm}, 80\, \text{mm}, 120\, \text{mm}\}$
, focusing initially on direct distance. To evaluate the on-surface functionality, we perform $n=48$ on-surface measurements of different shapes. We analyze three planes with distances of \( l_{\text{gt, plane}} = \{40\, \text{mm}, 80\, \text{mm}, 120\, \text{mm}\} \), complemented by convex and concave shapes. For the convex and concave distances, the "Wave" shape is represented by $ l_{\text{gt, wave}} = 78.573\, \text{mm} $, the curve shape by $ l_{\text{gt, curve}} = 68.625\, \text{mm} $, and the triangle shape with $ l_{\text{gt, triangle}} = 65\, \text{mm} $. We compute the metrics for both the basic on-surface variant and the spline-interpolated variant.

\paragraph{Qualitative Evaluation.} A critical aspect of this method is the ability to perform online measurements with surgical tools within the camera scene. To evaluate this functionality, we conduct several measurements using real tools from the da Vinci Surgical System on a realistic phantom representing a hepatectomy (liver resection) (\Cref{fig:environments}). The surgical phantom used in the evaluation is provided by the Experimental Surgery Department of Charité Universitätsmedizin Berlin \cite{remde_xr-developer_2023}. Since the phantom does not have exact distances to verify, we limit our evaluation to a qualitative assessment of the tooltip estimation and point cloud.

The third environment is based on a video recording from an actual surgical procedure \cite{ye_self-supervised_2017} (\Cref{fig:environments}). We utilize a publicly available dataset. It contains rectified stereo images captured during real robotic surgery by the Hamlyn Centre and was published in conjunction with \citet{ye_self-supervised_2017}. This dataset does not include ground truth data, therefore our evaluation is limited to qualitative analysis of the effects of smoke, blood, and reflections on the reconstructed surface. We sample individual stereo images from the dataset and apply the proposed method to these images, subsequently visualizing the outcomes.

\section{Results}
\label{sec:results}

Below, we present the comprehensive results of our quantitative and qualitative experiments.

\subsection{Quantitative Evaluation in the Playground}

\paragraph{Direct measurements.} 
\Cref{table:directmeasurements} presents the measurement results for the three different direct distances. With RAFT Stereo as a component for disparity estimation, the MAE for all three measurements is below $0.3$ mm in the offline setting, demonstrating high accuracy. The STD of the error remains under $0.2$ mm, indicating a robust performance. With the maximum error being under 0.8 mm, the results prove to be highly accurate. In the online setting, the results are less accurate than the measurement set. This seems to be plausible as it is difficult to handle instruments as precisely.

In contrast, accurate distance measurements using the SGBM algorithm for disparity estimation are not feasible, as shown in \Cref{table:directmeasurements}. Although we observed isolated precise measurements, the SGBM algorithm is not able to produce accurate results over multiple measurements. 

In summary, both the textureless regions and the poor lighting conditions in the images present a unique challenge for disparity estimation, as clearly demonstrated by the SGBM baseline example. From this point forward, we solely consider RAFT-Stereo for disparity estimation, as SGBM led to implausible values.

\begin{table}[ht]
	\centering
	\resizebox{\columnwidth}{!}{
		\begin{tabular}{l|c|c|c|c||c|c}
			\toprule
			\multirow{2}{*}{\textbf{Measurement}} & \multicolumn{4}{c||}{\textbf{Offline}} & \multicolumn{2}{c}{\textbf{Online}} \\
			\cmidrule(lr){2-5} \cmidrule(lr){6-7}
			& \multicolumn{2}{c|}{MAE $\pm$ STD} & \multicolumn{2}{c||}{max} & MAE $\pm$ STD & max \\
			\cmidrule(lr){2-3} \cmidrule(lr){4-5} \cmidrule(lr){6-6} \cmidrule(lr){7-7}
            & SGBM & RAFT & SGBM & RAFT & RAFT & RAFT \\
			\midrule
			40 mm   & 732.97 $\pm$ 500.94 & 0.18 $\pm$ 0.15 & 1362.22 & 0.67 & 0.98 $\pm$ 0.44 & 1.58 \\
            80 mm   & 615.34 $\pm$ 495.42 & 0.28 $\pm$ 0.19 & 1241.64 & 0.79 & 1.52 $\pm$ 1.00 & 2.59 \\
            120 mm  & 618.96 $\pm$ 491.09 & 0.24 $\pm$ 0.18 & 1311.19 & 0.67 & 0.76 $\pm$ 0.47 & 1.51 \\
			\bottomrule
		\end{tabular}
    }
	\caption{Comparison of SGBM and RAFT Euclidean measurements with offline and online selection.}
	\label{table:directmeasurements}
\end{table}

\begin{figure}[ht]
    \centering
    \includegraphics[width=.98\columnwidth]{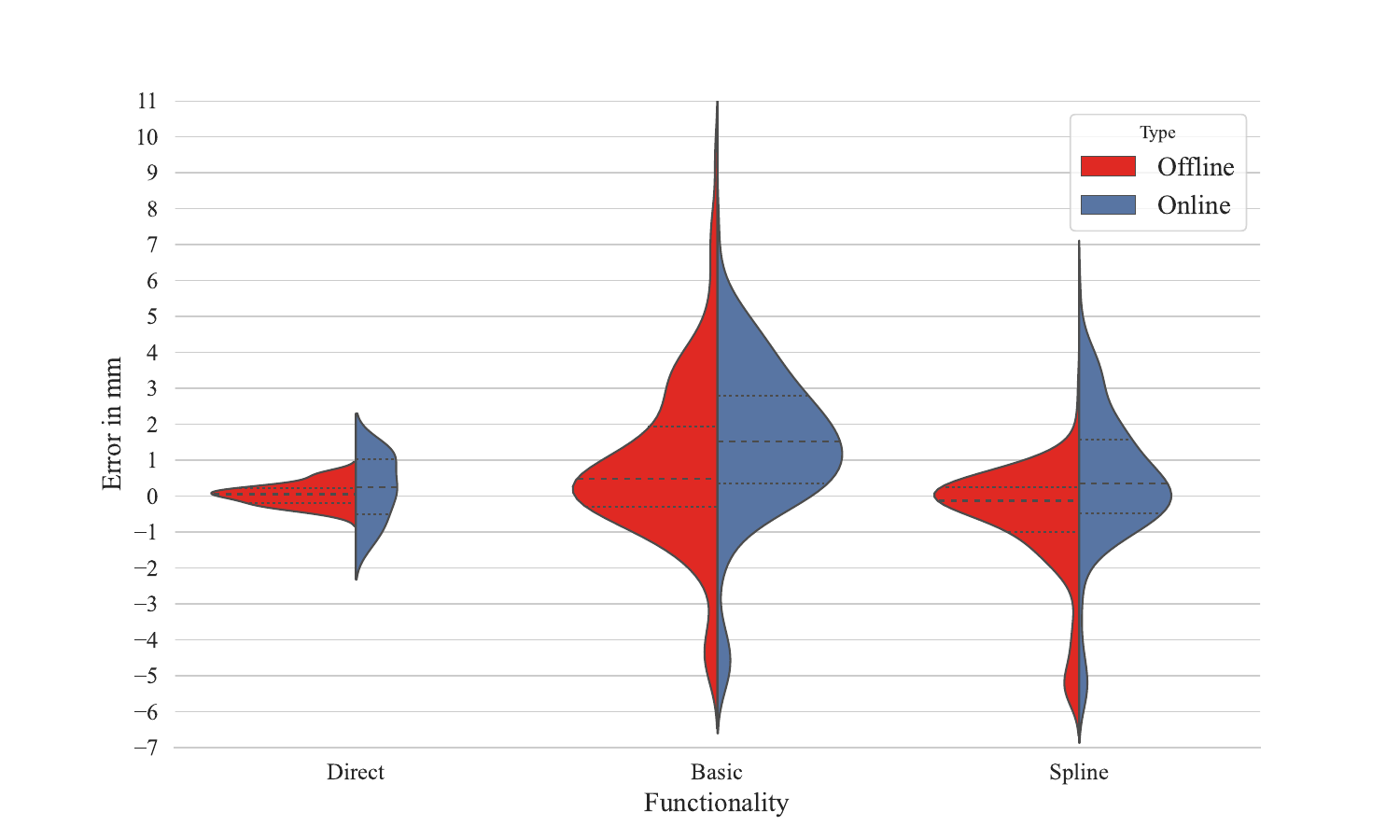}
    \caption{Measurement error in mm for direct, basic, and spline measurements in the online and offline setting.}
    \label{fig:measurements}
\end{figure}
\begin{figure}[ht]%
\centering
    \begin{tabular}{p{2.2 cm} p{2.2 cm} p{2.2 cm}}
        \includegraphics[width=2.2 cm]{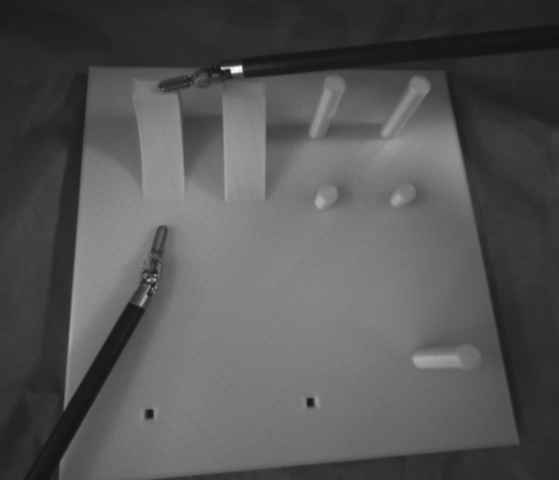} &
        \includegraphics[width=2.2 cm]{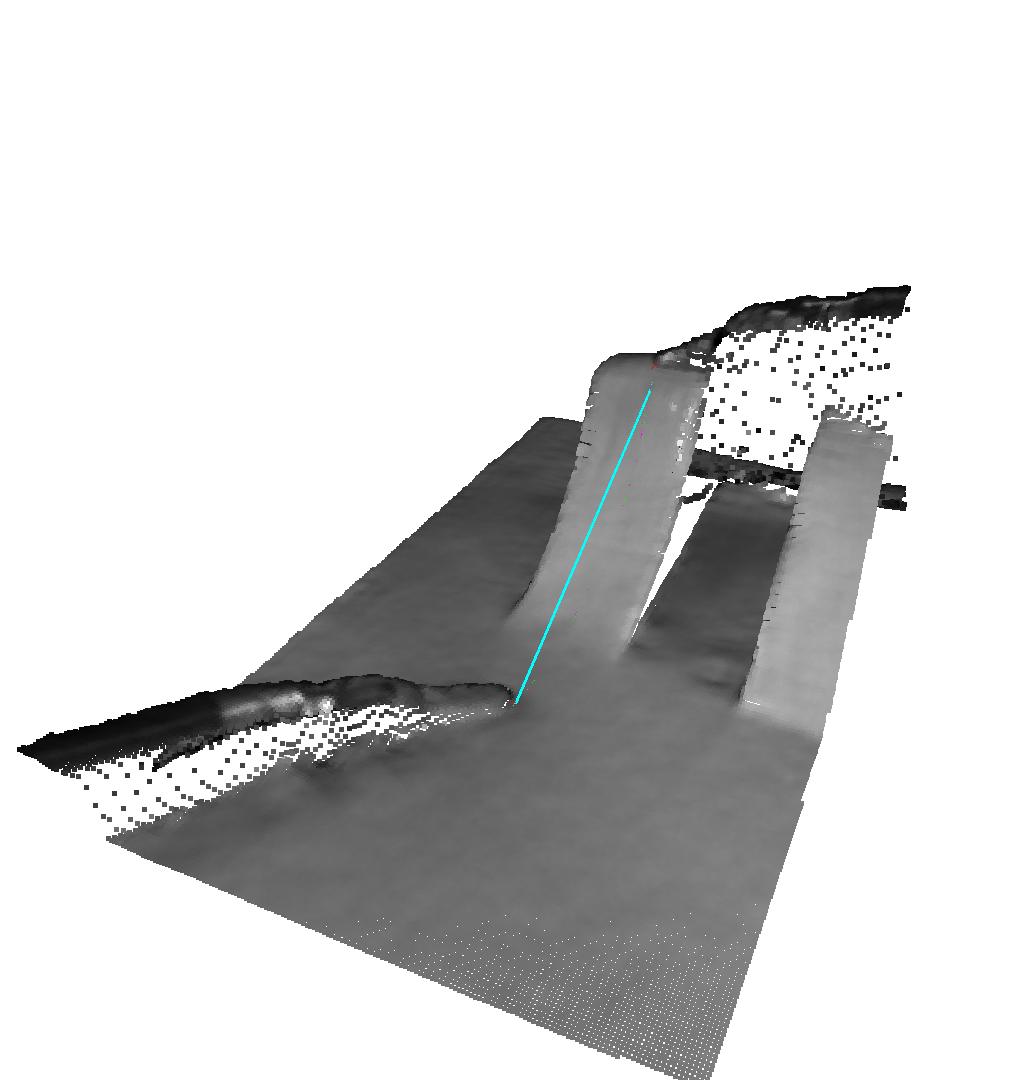} &
        \includegraphics[width=2.2 cm]{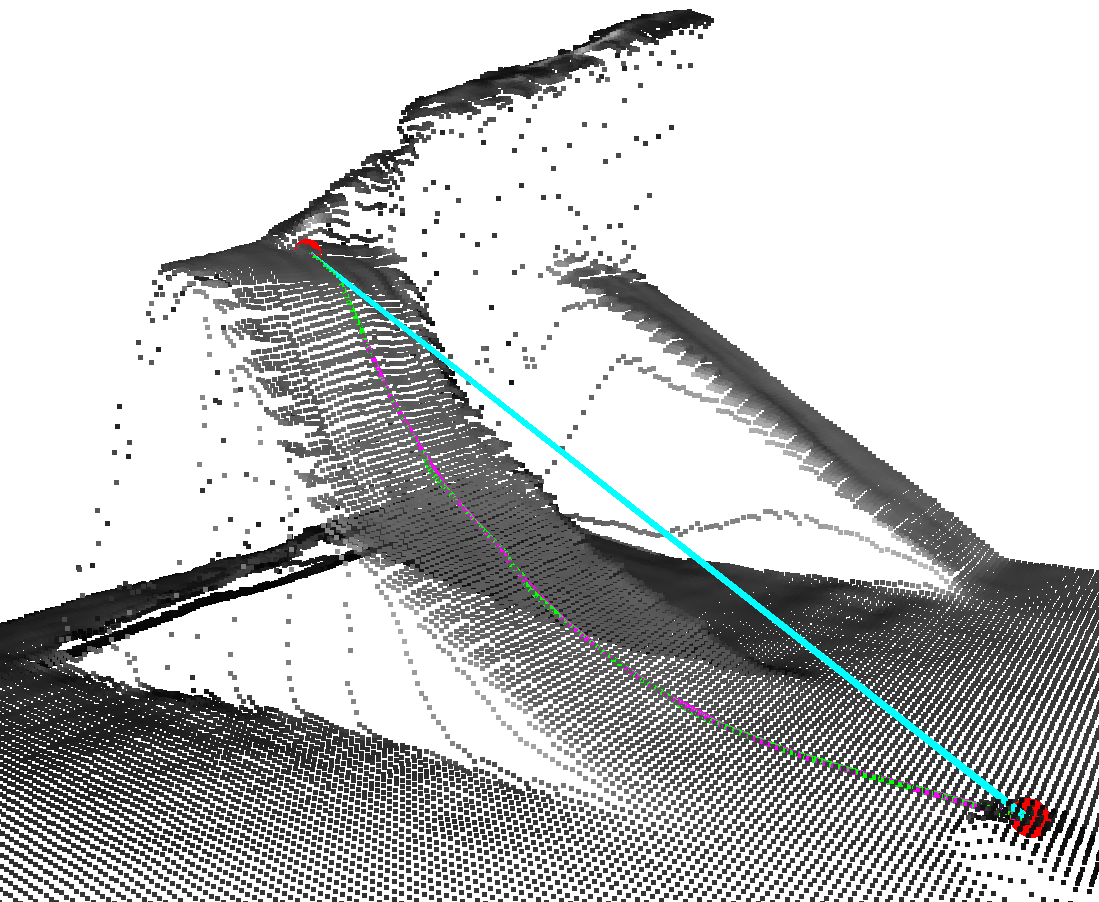} \\        
    \end{tabular}
    \caption{Exemplary online measurement in the Playground.}
    \label{fig:playground}
\end{figure}

\paragraph{On-surface measurements.} \Cref{table:onsurface} shows the results for the on-surface measurements. First, we can observe that on-surface measurements are less accurate than direct measurements. The spline interpolation is able to reduce measurement errors (\Cref{fig:measurements}), but they are still present. The errors may come from the higher complexity as every deviation from the real world to the point cloud is summed up but our analysis reveals an even more significant factor. While the point cloud itself is highly accurate, as evidenced by the direct measurements in (\Cref{fig:measurements}) and the point cloud in \Cref{fig:playground}, the bottleneck lies in the mesh creation. The PSR is not able to reproduce the correct surface over all measurement samples.

\begin{table}[ht]
	\centering
	\resizebox{\columnwidth}{!}{
		\begin{tabular}{l|c|c|c|c||c|c|c|c}
			\toprule
			\multirow{2}{*}{\textbf{Measurement}} & \multicolumn{4}{c||}{\textbf{Offline}} & \multicolumn{4}{c}{\textbf{Online}} \\
			\cmidrule(lr){2-5} \cmidrule(lr){6-9}
			& \multicolumn{2}{c|}{MAE $\pm$ STD} & \multicolumn{2}{c||}{max} & \multicolumn{2}{c|}{MAE $\pm$ STD} & \multicolumn{2}{c}{max} \\
			\cmidrule(lr){2-3} \cmidrule(lr){4-5} \cmidrule(lr){6-7} \cmidrule(lr){8-9}
            & Basic & Spline & Basic & Spline & Basic & Spline & Basic & Spline \\
			\midrule
			40 mm     & 0.83 $\pm$ 0.98 & 0.24 $\pm$ 0.18 & 4.26 & 0.66 & 2.08 $\pm$ 1.47 & 1.63 $\pm$ 1.07 & 5.56 & 4.00 \\
            80 mm     & 1.50 $\pm$ 1.83 & 0.39 $\pm$ 0.33 & 8.05 & 1.38 & 2.72 $\pm$ 1.54 & 2.11 $\pm$ 1.40 & 5.48 & 4.52 \\
            120 mm    & 2.05 $\pm$ 2.49 & 0.47 $\pm$ 0.44 & 9.60 & 2.28 & 1.53 $\pm$ 1.28 & 1.11 $\pm$ 0.94 & 6.35 & 4.97 \\
            Wave      & 2.68 $\pm$ 1.78 & 3.26 $\pm$ 1.87 & 7.21 & 5.67 & 2.82 $\pm$ 1.35 & 0.74 $\pm$ 0.43 & 5.32 & 1.58 \\
            Curve     & 1.71 $\pm$ 1.20 & 1.19 $\pm$ 0.57 & 4.40 & 2.20 & 2.22 $\pm$ 2.03 & 2.30 $\pm$ 2.02 & 7.54 & 5.88 \\
            Triangle  & 1.35 $\pm$ 1.34 & 0.47 $\pm$ 0.31 & 3.78 & 1.06 & 1.12 $\pm$ 0.62 & 0.51 $\pm$ 0.48 & 2.47 & 3.43 \\
			\bottomrule
		\end{tabular}
    }
	\caption{Comparison of Results from basic and spline-interpolated on-surface measurements with offline and online selection.}
	\label{table:onsurface}
\end{table}

\subsection{Qualitative Evaluation}

\begin{figure}[ht]%
\centering
    \begin{tabular}{p{3.5cm} p{3cm}}
        \includegraphics[width=3.5cm]{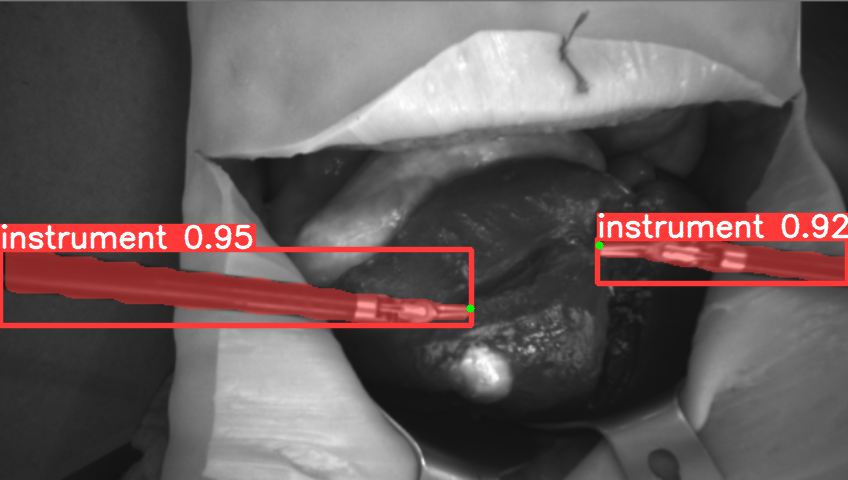} &
        \includegraphics[width=3cm]{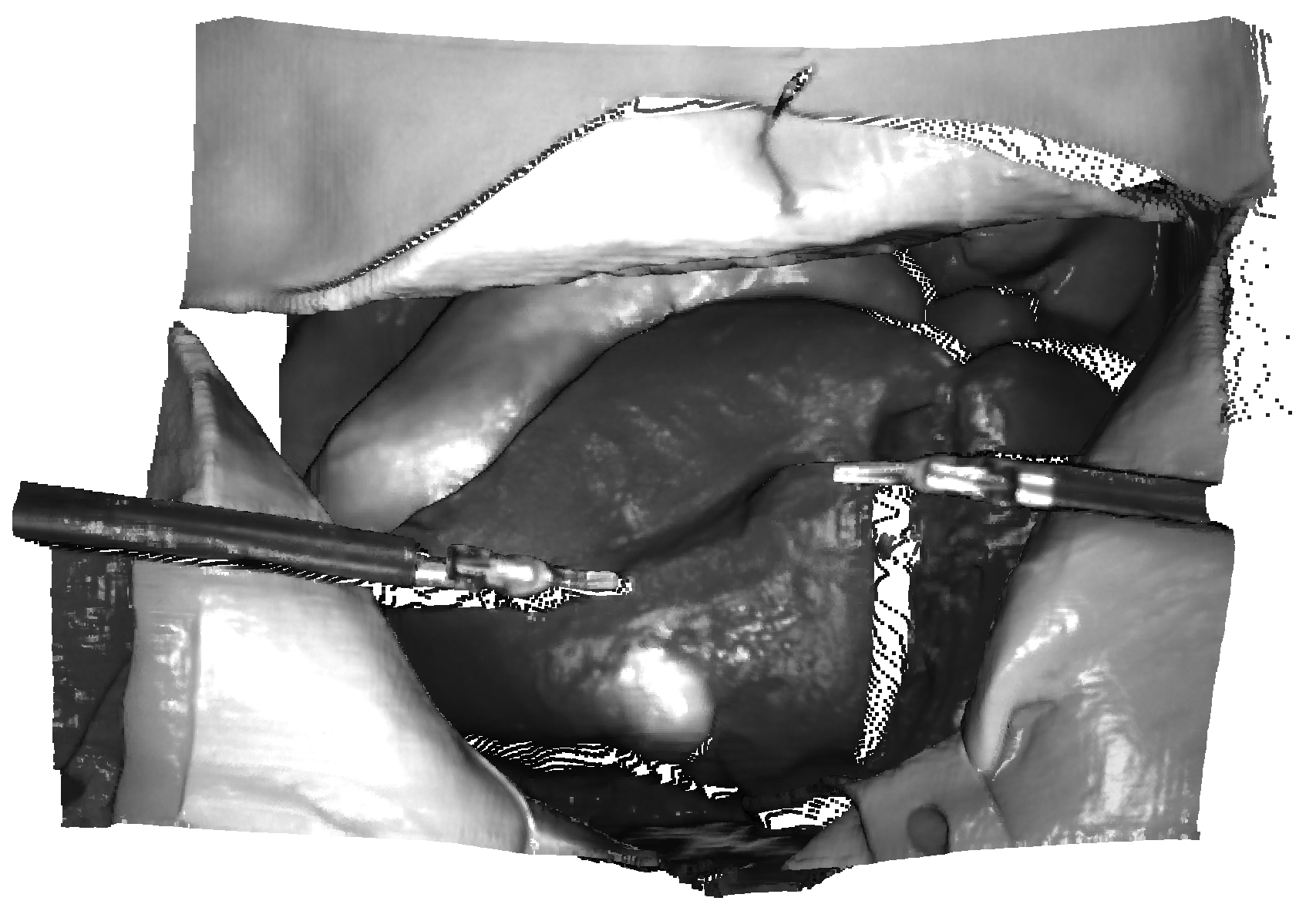} \\
        \centering
        \small Original left image. &
        \centering
        \small Point cloud. \\
        
    \end{tabular}
    \caption{Exemplary reconstruction of the Phantom and instruments.}
    \label{fig:phantom}
\end{figure}

\Cref{fig:phantom} shows an exemplary measurement of the surgical phantom using two graspers from the da Vinci Surgical System. The left image shows the segmented instruments and the corresponding estimated tips in green. The right image shows the point cloud. Overall both, the tooltip estimation and the re-projection are highly accurate.

\begin{figure}[ht]%
\centering
    \begin{tabular}{p{3 cm} p{2.7cm}}
        \includegraphics[width=3cm]{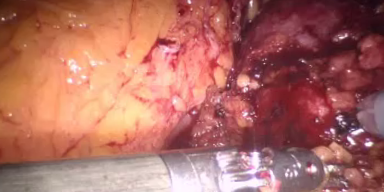} &
        \includegraphics[width=2.7cm]{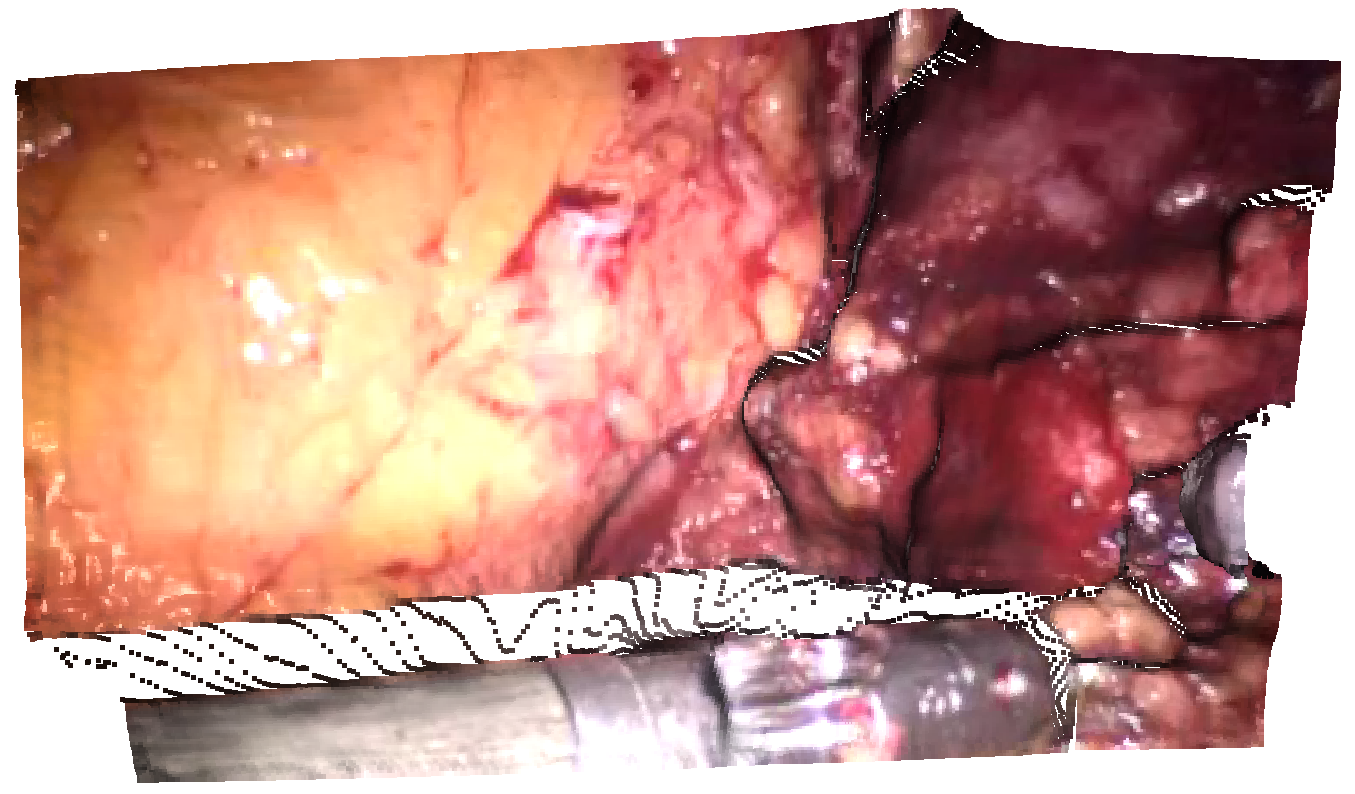} \\
        \includegraphics[width=3cm]{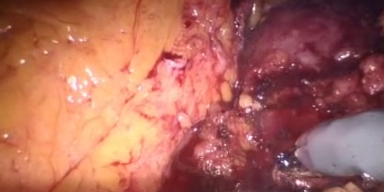} &
        \includegraphics[width=2.7cm]{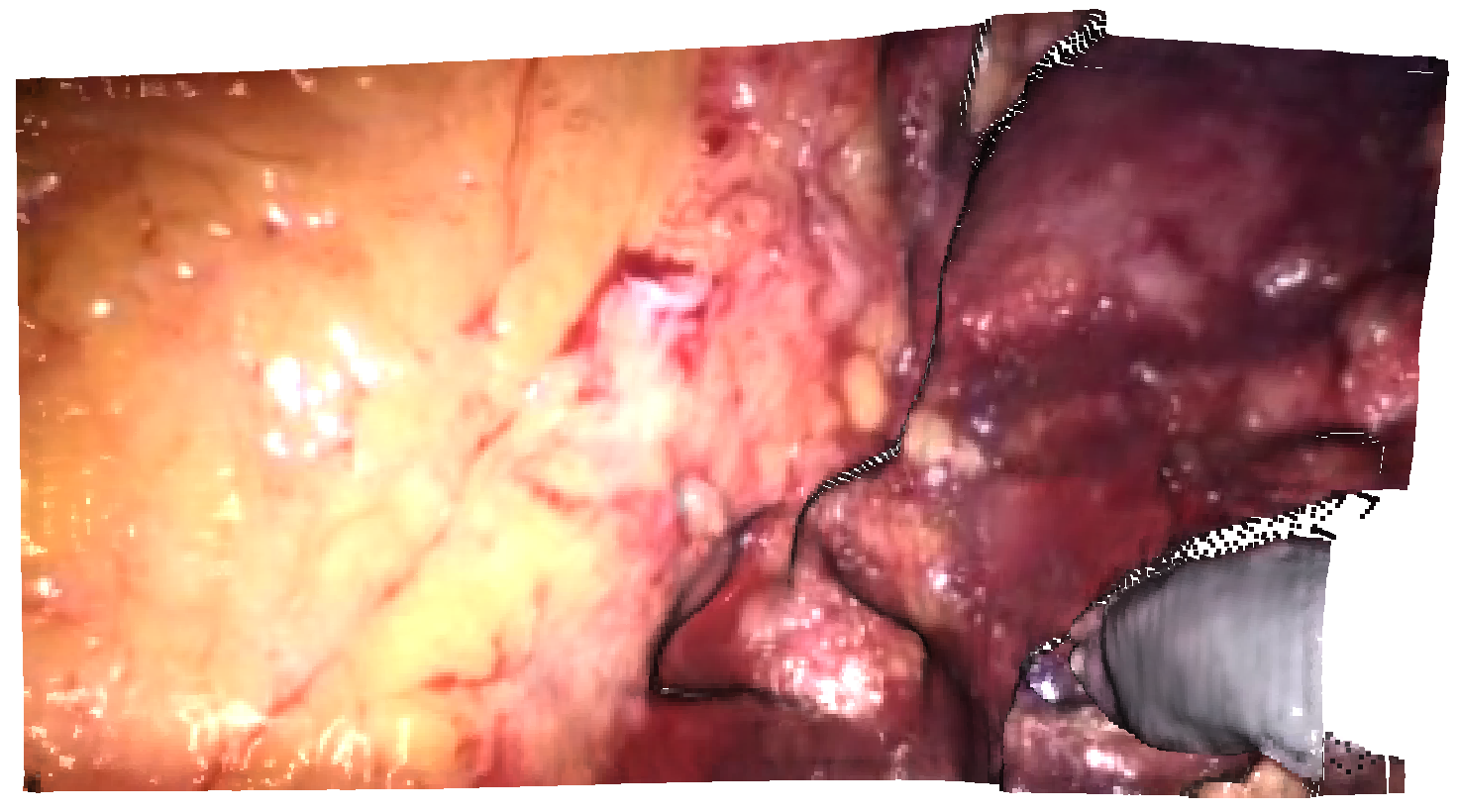} \\
        \includegraphics[width=3cm]{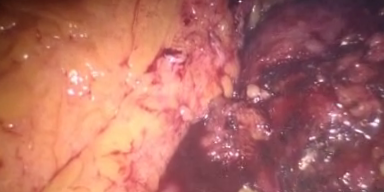} &
        \includegraphics[width=2.7cm]{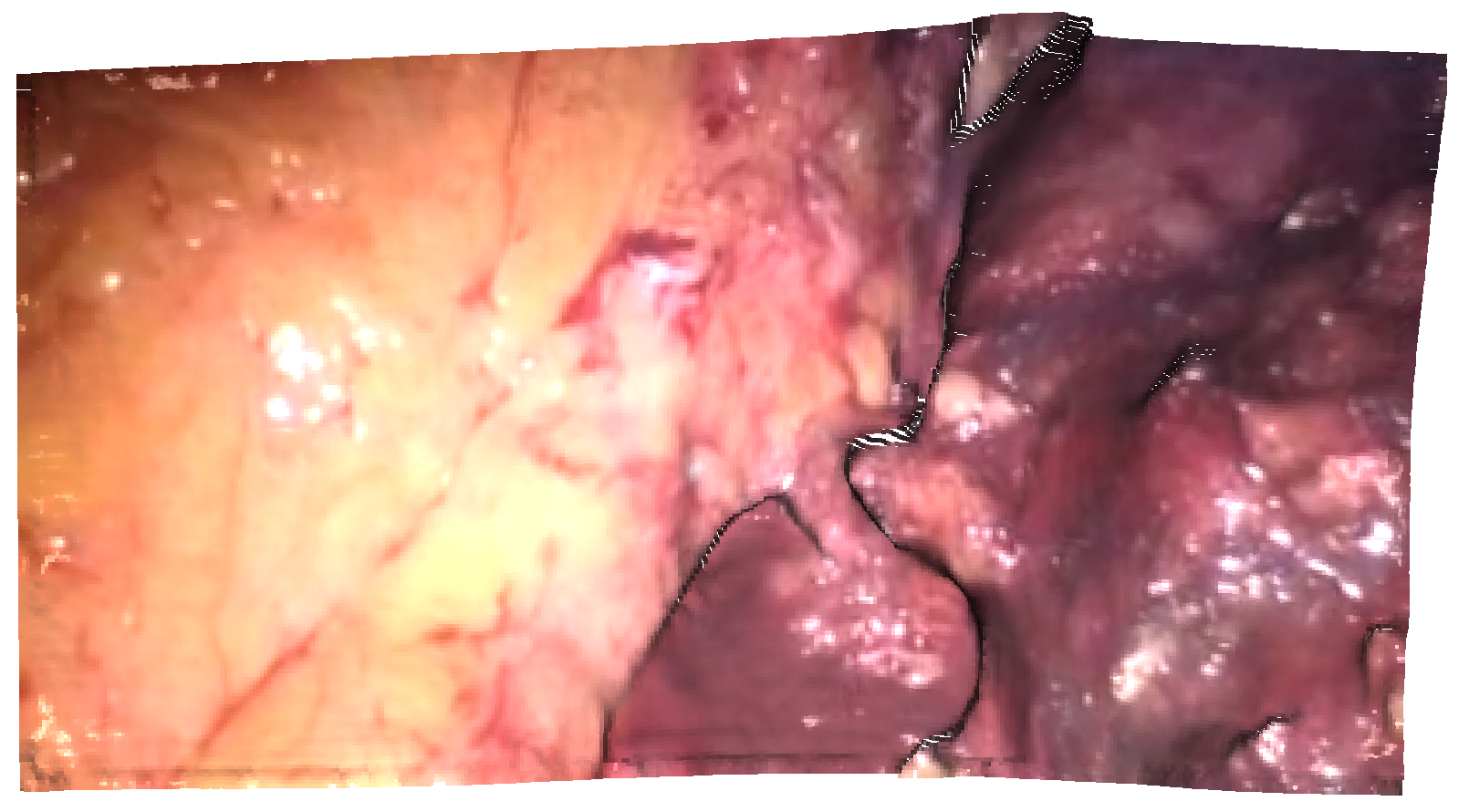}\\
        \centering
        \small Original left image. &
        \centering
        \small Point cloud. \\
        
    \end{tabular}
    \caption{\centering Exemplary point clouds of real surgery images.}
    \label{fig:realsurgery}
\end{figure}

In addition, we perform a qualitative assessment of the applicability of our framework in a real environment by applying it to actual images obtained from the da Vinci Surgical System. \Cref{fig:realsurgery} shows exemplary surface reconstructions for these real-world images. The results are highly promising, with no noticeable noise. In addition, the framework demonstrates robustness in handling reflections and the presence of blood or smoke, which do not pose any issues.

\section{Discussion}
\label{sec:Discussion}

As the current implementation of the method is only a prototype, there are some limitations to be aware of.

\subsection{Limitations}

Intraoperative online measurements require a real-time version of our method, making YOLOv8 particularly suitable for this purpose. Although RAFT-Stereo and similar techniques can achieve real-time performance, they require significantly more computational resources due to their reliance on GRUs. Currently, the primary bottleneck in our system is mesh generation using PSR. Despite the lack of graphics processing unit acceleration, this step remains the most critical aspect of our prototype implementation. Even with minimal noise and high accuracy of point clouds, on-surface measurements occasionally produce outliers due to imperfections in the resulting mesh. Our method offers several advantages for surgical procedures, primarily by meeting the key requirements (R1-R8) that emphasise minimal disruption to the medical workflow. However, predicting the adoption of such a method by surgeons in general is challenging \cite{hemmer2022factors}. This is exacerbated by the lack of a comparable system and data on usage frequency. An additional point of consideration is the optimal measurement process in practical settings. Currently, accurate measurements require that both surgical instruments, including their tooltips, remain visible within the camera's field of view.

\subsection{Path to Deployment}

Addressing the limitations mentioned above, several research directions can enhance our method. First, it is worth considering the enhancement of the mesh creation component and optimizing hyperparameters. Exploring AI models for generating a mesh from point clouds is promising \cite{Miyauchi2022}. Similar to RAFT-Stereo, the AI model could recognize structures and thus represent a more intelligent approach than PSR. Our method's modularity allows for the optimization or replacement of elements in future studies. Second, it would be of interest to refine the prototype by using a laparoscopic stereo camera and involving surgeons in real-world measurements \cite{Dowrick2023}. In addition, we consider methodologies that map multiple reconstructions of internal structures, allowing for persistent reference points even when they transition out of the camera's field of view. Collaborating with surgeons will pinpoint weaknesses and possible applications. Additionally, integrating workflow recognition could enable automatic measurement of actions like cuts. Lastly, as reliable measurement accuracy is crucial, integrating measurement uncertainties by estimating error margins is an important future research avenue. 

\section{Conclusion}

This work aims to develop a versatile measurement tool for laparoscopy. We outline the fundamental characteristics of laparoscopy and establish the key requirements for a laparoscopic measurement tool. Based on these requirements, we devise a method that intelligently integrates the results of existing research into a formally defined framework.

We evaluate our method through a series of experiments designed to test various functionalities. To achieve this, we implement the method as a prototype, integrating several state-of-the-art components and establishing appropriate experimental environments. For distance measurements, we design and 3D print a CAD model, while a surgical phantom serves for more realistic application scenarios. We construct datasets for both environments, opting for a challenging setting to evaluate the robustness of our method. In addition, we supplement these two datasets with a third public dataset comprising real-world laparoscopic images.

The results show the potential of our proposed human-AI-based measurement method. In addition to exhibiting a remarkably high accuracy, it demonstrates substantial robustness when applied to real-world image data. The inclusion of AI-based components, such as RAFT-Stereo and YOLOv8, is instrumental in achieving these results. Moreover, we successfully show the essential functionality of online measurements by employing surgical instruments to establish reference points. Although the measurement outcomes are not as accurate as those in the offline selection, they still indicate potential. Considering that we do not optimize the individual components for laparoscopic application in the current setting, these results serve as a proof of concept. 

\bibliography{aaai24}

\end{document}